\documentclass[journal, onecolumn, final]{IEEEtran}
\usepackage{array}
\usepackage[numbers]{natbib}
\usepackage{amssymb,amsmath,subcaption}
\usepackage{graphicx}\graphicspath{{figs/}{eps/}}
\usepackage{psfrag}
\usepackage{ulem}
\usepackage{algorithmic}
\usepackage{algorithm}
\usepackage{varioref}
\usepackage{booktabs}
\usepackage{tabularx}
\usepackage[para,online,flushleft]{threeparttable}

\newcommand{\mat}[1]{\left[\begin{array}{#1}}

\usepackage{url}
\begin{document}
\title{Learning a hyperplane classifier by minimizing an exact bound on the VC dimension}

\author{Jayadeva\\
Department of Electrical Engineering, Indian Institute of Technology, Delhi, Hauz Khas, New Delhi - 110016, INDIA. e-mail: jayadeva@ee.iitd.ac.in
\thanks{Accepted Author Manuscript (Neurocomputing, Elsevier). For queries regarding commercial licensing of the MCM and its variants, please contact the Foundation for Innovation and Technology Transfer, IIT Delhi.}}

\maketitle


\begin{abstract}
The VC dimension measures the complexity of a learning machine, and a low VC dimension leads to good generalization. While SVMs produce state-of-the-art learning performance, it is well known that the VC dimension of a SVM can be unbounded; despite good results in practice, there is no guarantee of good generalization. In this paper, we show how to learn a hyperplane classifier by minimizing an exact, or \boldmath{$\Theta$} bound on its VC dimension. The proposed approach, termed as the Minimal Complexity Machine (MCM), involves solving a simple linear programming problem. Experimental results show, that on a number of benchmark datasets, the proposed approach learns classifiers with error rates much less than conventional SVMs, while often using fewer support vectors. On many benchmark datasets, the number of support vectors is less than one-tenth the number used by SVMs, indicating that the MCM does indeed learn simpler representations.
\end{abstract}


\section{Introduction}

Support vector machines are amongst the most widely used machine learning techniques today. The classical SVM \citep{L1svm} has evolved into a multitude of diverse formulations with different properties. The most commonly used variants are the maximum margin $L_1$ norm SVM \citep{L1svm}, and the least squares SVM (LSSVM) \citep{suykens1999least}, both of which require the solution of a quadratic programming problem. In the last few years, SVMs have been applied to a number of applications to obtain cutting edge performance; novel uses have also been devised, where their utility has been amply demonstrated \citep{Yang201477, Ding20143, Wu201498, Li201331, Wu2014119, Wenjian2013116, Peng2013134, Jalalian2013270, Peng2013486, Huang2013118, Gomes20123, Ni2014127, Peng2013197, Yin2014224, LópezChau2013198, Kumar2014271, Ruano2014273, Liu201397, Zhao2013225, Ji2014281, Zhao20121, Sun2014111}. SVMs were motivated by the celebrated work of Vapnik and his colleagues on generalization, and the complexity of learning. It is well known the capacity of a learning machine can be measured by its Vapnik-Chervonenkis (VC) dimension. The VC dimension can be used to estimate a probabilistic upper bound on the test set error of a classifier. A small VC dimension leads to good generalization and low error rates on test data.\\

In his widely read tutorial, Burges \citep{burges1998} states that SVMs can have a very large VC dimension, and that \textit{``at present there exists no theory which shows that good generalization performance is guaranteed for SVMs''}. This paper shows how to learn a classifier with large margin, by minimizing an exact (\boldmath{$\Theta$}) bound on the VC dimension. In other words, the proposed objective linearly bounds the VC dimension from both above and below. We show that this leads to a simple linear programming problem. This approach is generic, and it suggests numerous variants that can be derived from it - as has been done for SVMs. Experimental results provided in the sequel show that the proposed Minimal Complexity Machine outperforms conventional SVMs in terms of test set accuracy, while often using far fewer support vectors. That the approach minimizes the machine capacity may be guaged from the fact that on many datasets, the MCM yields better test set accuracy while using less than $\frac{1}{10}$-th the number of support vectors obtained by SVMs.

The motivation for the MCM originates from some sterling work on generalization \citep{shawe1996framework, shawetaylor98, vapnik98, scholkopf2002learning}. We restrict our attention in this paper to a given binary classification dataset for which a hyperplane classifier needs to be learnt. Consider such a binary classification problem with data points $x^i, i = 1, 2, ..., M$, and where samples of class +1 and -1 are associated with labels $y_i = 1$ and $y_i = -1$, respectively. We assume that the dimension of the input samples is $n$, i.e. $x^i = (x_1^i, x_2^i, ..., x_n^i)^T$. For the set of all gap tolerant hyperplane classifiers with margin $d \geq d_{min}$, Vapnik \citep{vapnik98} showed that the VC dimension $\gamma$ is bounded by
\begin{equation}\label{eqnh}
\gamma \leq 1 + \operatorname{Min}(\frac{R^2}{d^2_{min}}, n)
\end{equation}
where $R$ denotes the radius of the smallest sphere enclosing all the training samples. Burges, in \citep{burges1998}, stated that \textit{``the above arguments strongly suggest that algorithms that minimize $\frac{R^2}{d^2}$ can be expected to give better generalization performance. Further evidence for this is found in the following theorem of (Vapnik, 1998), which we quote without proof''}. We follow this line of argument and show, through a constructive result, that this is indeed the case.\\

The remainder of this paper is organized as follows. Section \ref{linmcm} outlines the proposed optimization problem for a linear hyperplane classfier in the input space. Section \ref{kmcm} discusses the extension of the Minimum Complexity Machine to the kernel case. Section \ref{experimental} is devoted to a discussion of results obtained on selected benchmark datasets. Section \ref{conclusion} contains concluding remarks. In Appendix \ref{appendix1}, we derive an exact bound for the VC dimension of a hyperplane classifier. Appendix \ref{appendix2} deals with the formulation of the hard margin MCM.

\section{The Linear Minimal Complexity Machine} \label{linmcm}
We first consider the case of a linearly separable dataset. By definition, there exists a hyperplane that can classify these points with zero error. Let the separating hyperplane be given by
\begin{equation}
 u^Tx + v = 0.
\end{equation}

Let us denote
\begin{gather}
 h = \frac{\operatorname*{Max}_{i = 1, 2, ..., M} \|u^T x^i + v\|}{\operatorname*{Min}_{i = 1, 2, ..., M} \|u^T x^i + v\|}.
\end{gather}
In Appendix \ref{appendix1}, we show that $h$ may also be written as
\begin{gather}
 h = \frac{\operatorname*{Max}_{i = 1, 2, ..., M} \;y_i(u^T x^i + v)}{\operatorname*{Min}_{i = 1, 2, ..., M} \;y_i(u^T x^i + v)},
\end{gather}
and we show that there exist constants $\alpha, \beta > 0$, $\alpha, \beta \in \Re$ such that
\begin{equation}\label{exactbound}
 \alpha h^2 \leq \gamma \leq \beta h^2,
\end{equation}
or, in other words, $h^2$ constitutes a tight or exact ($\theta$) bound on the VC dimension $\gamma$. An exact bound implies that $h^2$ and $\gamma$ are close to each other.\\

 Therefore, the machine capacity can be minimized by keeping $h^2$ as small as possible. Since the square function $(\cdot)^2$ is monotonically increasing, we can minimize $h$ instead of $h^2$. We now formulate an optimization problem that tries to find the classifier with smallest machine capacity that classifies all training points of the linearly separable dataset correctly; this problem is given by

\begin{equation}\label{minh1}
\operatorname*{Minimize  }_{u, v} \; h ~=~ \frac{\operatorname*{Max}_{i = 1, 2, ..., M} \; y_i(u^T x^i + v)}{\operatorname*{Min}_{i = 1, 2, ..., M} \; y_i(u^T x^i + v)}
\end{equation}

Note that in deriving the exact bound in Appendix \ref{appendix1}, we assumed that the separating hyperplane $u^Tx + v = 0$ correctly separates the linearly separable training points; consequently, no other constraints are present in the optimization problem (\ref{minh1}).\\

In Appendix \ref{appendix2}, we show that the optimization problem (\ref{minh1}) may be reduced to the problem
\begin{gather}
\operatorname*{Min}_{w, b, h} ~~h \label{objm4}\\
h \geq y_i \cdot [{w^T x^i + b}], ~i = 1, 2, ..., M \label{consm41}\\
y_i \cdot [{w^T x^i + b}] \geq 1, ~i = 1, 2, ..., M, \label{consm42}
\end{gather}
where $w \in \Re^n$, and $b, h \in \Re$. We refer to the problem (\ref{objm4}) - (\ref{consm42}) as the hard margin Linear Minimum Complexity Machine (Linear MCM). \\

Note that the variable $h$ in (\ref{objm4}) and that in (\ref{exactbound}) refer to the same functional. By minimizing $h$ in (\ref{objm4}), we are minimizing an exact bound on $\gamma$, the VC dimension of the classifier. Once $w$ and $b$ have been determined by solving (\ref{objm4})-(\ref{consm42}), the class of a test sample $x$ may be determined from the sign of the discriminant function
\begin{equation}\label{testresult}
 f(x) = w^T x + b
\end{equation}

In general, datasets will not be linearly separable. The soft margin equivalent of the MCM is obtained by introducing additional slack variables, and is given by
\begin{gather}
\operatorname*{Min}_{w, b, h} ~~h + C \cdot \sum_{i = 1}^M q_i \label{obj5}\\
h \geq y_i \cdot [{w^T x^i + b}] + q_i, ~i = 1, 2, ..., M \label{cons51}\\
y_i \cdot [{w^T x^i + b}] + q_i \geq 1, ~i = 1, 2, ..., M \label{cons52} \\
q_i \geq 0, ~i = 1, 2, ..., M. \label{cons53}
\end{gather}
Here, the choice of $C$ allows a tradeoff between the complexity (machine capacity) of the classifier and the classification error.\\

Once $w$ and $b$ have been determined, the class of a test sample $x$ may be determined as before by using the sign of $f(x)$ in (\ref{testresult}). In the sequel, we show how to extend the idea to nonlinearly separable datasets.

\section{The Kernel MCM}\label{kmcm}
We consider a map $\phi(x)$ that maps the input samples from $\Re^n$ to $\Re^l$, where $l > n$. The separating hyperplane in the image space is  given by

\begin{equation}
 u^T \phi(x) + v = 0.
\end{equation}

Following (\ref{obj5}) - (\ref{cons52}), the corresponding optimization problem for the kernel MCM may be shown to be

\begin{gather}
\operatorname*{Min}_{w, b, h, q} \; h + C \cdot \sum_{i = 1}^M q_i \label{objk6}\\
h \geq y_i \cdot [{w^T \phi(x^i) + b}] + q_i, ~i = 1, 2, ..., M \\
y_i \cdot [{w^T \phi(x^i) + b}] + q_i \geq 1, ~i = 1, 2, ..., M \label{consk61} \\
q_i \geq 0, ~i = 1, 2, ..., M.
\end{gather}

The image vectors $\phi(x^i), i = 1, 2, ..., M$ form an overcomplete basis in the empirical feature space, in which $w$ also lies. Hence, we can write
\begin{equation}\label{weqsumlambda}
 w = \sum_{j = 1}^M \lambda_j \phi(x^j).
\end{equation}

Therefore,
\begin{gather}\label{simpliphi}
 w^T \phi(x^i) + b = \sum_{j = 1}^M \lambda_j \phi(x^j)^T\phi(x^i) + b = \sum_{j = 1}^M \lambda_j K(x^i, x^j) + b,
\end{gather}

where $K(p, q)$ denotes the Kernel function with input vectors $p$ and $q$, and is defined as
\begin{equation}
 K(p, q) = \phi(p)^T \phi(q).
\end{equation}\label{kernel}

Substituting from (\ref{simpliphi}) into (\ref{objk6}) - (\ref{consk61}), we obtain the following optimization problem.
\begin{gather}
\operatorname*{Min}_{w, b, h, q} \; h + C \cdot \sum_{i = 1}^M q_i \label{objk7}\\
h \geq y_i \cdot [\sum_{j = 1}^M \lambda_j K(x^i, x^j) + b] + q_i, ~i = 1, 2, ..., M\\
y_i \cdot [\sum_{j = 1}^M \lambda_j K(x^i, x^j) + b] + q_i \geq 1, ~i = 1, 2, ..., M \\ \label{consk71}
q_i \geq 0, ~i = 1, 2, ..., M.
\end{gather}

Once the variables $\lambda_j, j = 1, 2, ..., M$ and $b$ are obtained, the class of a test point $x$ can be determined by evaluating the sign of
\begin{equation}
f(x) ~=~ w^T \phi(x) + b ~=~ \sum_{j = 1}^M \lambda_j K(x, x^j) + b.
\end{equation}\label{testval}

Note that in (\ref{weqsumlambda}), the $\phi(x^j)$'s for which the corresponding $\lambda_j$'s are non-zero constitute the support of the vector $w$. Hence, $\phi(x^j)$'s for which the $\lambda_j$'s are non-zero may be termed as support vectors. The maximum number of support vectors is the number of training samples $M$, as in the case of the SVM.

\section{Experimental results}\label{experimental}
\begin{figure}[hbtp]
        \centering	
                \includegraphics[scale=0.5]{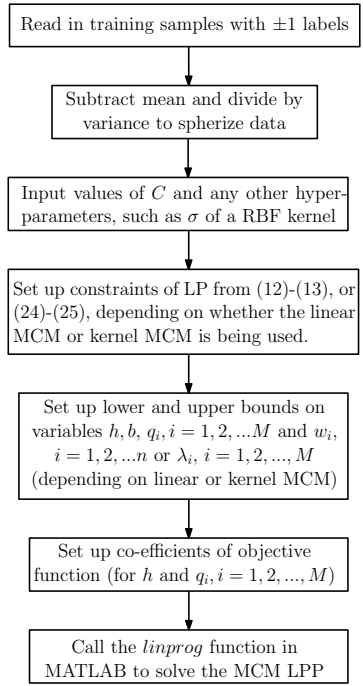}
                \caption{Illustration of the MATLAB flowchart for the MCM code.}
                \label{matlabmcm}
\end{figure}

The MCM was coded in MATLAB \citep{matlab}. Figure \ref{matlabmcm} provides a flowchart illustrating the MCM implementation; as may be seen, we use the \textit{linprog} function in MATLAB to solve the optimization problem comprising the MCM. 

In order to compare the MCM with SVMs, we chose a number of benchmark datasets from the UCI machine learning repository \citep{uciml}. Table \ref{tabledatasets} summarizes information about the number of samples and features of each dataset. Some of the benchmark datasets are multi-class ones. These have been learnt by adopting a one-versus-rest approach. Accuracies, CPU times, and the number of support vectors have been averaged across all the classifiers learnt in a one-versus-rest approach.

\begin{table}[htbp]
  \centering
\caption{Characteristics of the Benchmark Datasets used}
     \begin{tabular}{cc}
    \toprule
    dataset & Size (samples $\times$ features $\times$ classes) \\
    \midrule
          &  \\
    blogger & 100 $\times$ 6 $\times$ 2 \\
    fertility diagnosis & 100 $\times$ 9 $\times$ 2 \\
    promoters  & 106 $\times$ 57 $\times$ 2 \\
    echocardiogram  & 132 $\times$ 12 $\times$ 2 \\
    teaching assistant & 151 $\times$ 5 $\times$ 3 \\
    hepatitis  & 155 $\times$ 19 $\times$ 2 \\
    hayes & 160 $\times$ 5 $\times$ 3 \\
    plrx  & 182 $\times$ 12 $\times$ 2 \\
    seed  & 210 $\times$ 7 $\times$ 3 \\
    glass & 214 $\times$ 10 $\times$ 6 \\
    heartstatlog  & 270 $\times$ 13 $\times$ 2 \\
    horsecolic  & 300 $\times$ 27 $\times$ 2 \\
    haberman  & 306 $\times$ 3 $\times$ 2 \\
    ecoli & 336 $\times$ 8 $\times$ 3 \\
    house voters & 435 $\times$ 16 $\times$ 2 \\
    wholesale customer & 440 $\times$ 8 $\times$ 2 \\
    IPLD  & 583 $\times$ 10 $\times$ 2 \\
    balance & 625 $\times$ 4 $\times$ 3 \\
    australian  & 690 $\times$ 14 $\times$ 2 \\
    crx   & 690 $\times$ 15 $\times$ 2 \\
    transfusion  & 748 $\times$ 5 $\times$ 2 \\
    tic tac toe & 958 $\times$ 9 $\times$ 2 \\
    sorlie & 85 $\times$ 456 $\times$ 2 \\
    secom & 1567 $\times$ 591 $\times$ 2 \\
    tian   & 173 $\times$ 12626 $\times$ 2 \\
    \bottomrule
    \end{tabular}%
  \label{tabledatasets}%
\end{table}%

Table \ref{table1} summarizes five fold cross validation results of the linear MCM on the datasets listed in Table \ref{tabledatasets}. Accuracies are indicated as mean $\pm$ standard deviation, computed over the five folds. The table compares the linear MCM with the SVM using a linear kernel. Note that the MCM solves the primal problem and not the dual. The linear SVM is solved using the dual, by using a linear kernel. This is done because the SVM primal is numerically less attractive owing to the ill conditioning of the Hessian. Hence, there is a subtle difference between a linear MCM and a SVM (linear kernel). The values of $C$ were determined for the MCM by performing a grid search. The table also provides a comparison between the CPU times taken by the SVM and the MCM. For the implementation of the SVM, we use a small MATLAB code that solves the quadratic programming dual form of the classical $L_1$ norm SVM by calling the \textit{quadprog} command in MATLAB. This has been done to enable a comparison of the CPU times taken by the MCM and the SVM. Faster implementations based on active set methods are possible for the MCM; such methods exist for the SVM and include Platt's SMO \citep{plattsmo} and the 1SMO algorithm \citep{jdsumt2012}.

\begin{table}[htbp]
  \centering
  \caption{Linear MCM Results}
  \begin{threeparttable}
    \begin{tabular}{ccccc}
    \toprule
    datasets & \multicolumn{2}{c}{linear MCM}     & \multicolumn{2}{c}{linear SVM} \\
    \midrule
          & accuracy & time (s)  & accuracy & time (s) \\
    blogger & 69.00 $\pm$ 17.15 & 0.0012 $\pm$ 6.64e-5 & 58.00 $\pm$ 20.40 & 6.17 $\pm$ 2.51 \\
    fertility diagnosis & 88.00 $\pm$ 9.27 & 0.0013 $\pm$ 5.27e-5 & 86.00 $\pm$ 9.01 & 8.12 $\pm$ 1.47 \\
    promoters  & 74.08 $\pm$ 10.88 & 0.0014 $\pm$ 4.53e-5 & 67.78 $\pm$ 10.97 & 0.85 $\pm$ 0.03 \\
    echocardiogram  &  90.88 $\pm$ 5.75  & 0.0014 $\pm$ 4.62e-5 & 86.38 $\pm$ 4.50 & 0.72 $\pm$ 0.36 \\
    teaching assistant & 66.27 $\pm$ 6.77 & 0.0013 $\pm$ 3.46e-5 & 64.94 $\pm$ 6.56 & 16.07 $\pm$ 4.17 \\
    hepatitis  & 68.38 $\pm$ 6.26 & 0.0014 $\pm$ 3.82e-5 & 60.64 $\pm$ 7.19 & 1.90 $\pm$ 0.56 \\
    hayes & 76.32 $\pm$ 9.25 & 0.0012 $\pm$ 2.73e-5 & 73.56 $\pm$ 7.73 & 7.19 $\pm$ 3.81 \\
    plrx  & 71.83 $\pm$ 7.49 & 0.0015 $\pm$ 3.37e-5 & 71.42 $\pm$ 7.37 & 4.35 $\pm$ 0.78 \\
    seed  & 97.61 $\pm$ 1.51 & 0.0015 $\pm$ 3.97e-5 & 90.95 $\pm$ 4.09 & 12.37 $\pm$ 4.51 \\
    glass & 99.06 $\pm$ 1.16 & 0.0042 $\pm$ 5.56e-3 & 98.12 $\pm$ 1.75 & 11.83 $\pm$ 3.44 \\
    heartstatlog  & 84.81 $\pm$ 3.87 & 0.0018 $\pm$ 1.47e-5 & 82.59 $\pm$ 2.22 & 9.43 $\pm$ 4.25 \\
    horsecolic  & 81.00 $\pm$ 4.03 & 0.0021 $\pm$ 7.17e-5 & 80.26 $\pm$ 4.63 & 41.39 $\pm$ 13.93 \\
    haberman  & 73.89 $\pm$ 3.71 & 0.0019 $\pm$ 4.34e-5 & 72.56 $\pm$ 3.73 & 13.74 $\pm$ 6.63 \\
    ecoli & 96.73 $\pm$ 1.96 & 0.0023 $\pm$ 1.3e-4 & 96.73 $\pm$ 1.96 & 18.41 $\pm$ 2.57 \\
    house voters & 95.63 $\pm$ 1.84 & 0.0031 $\pm$ 1.87e-4 & 94.48 $\pm$ 2.46 & 15.77 $\pm$ 2.19 \\
    wholesale customer & 92.26 $\pm$ 1.97 & 0.0033 $\pm$ 1.07e4 & 91.13 $\pm$ 1.95 & 32.11 $\pm$ 8.29 \\
    IPLD  & 71.35 $\pm$ 2.93 & 0.0065 $\pm$ 4e-5 & 71.35 $\pm$ 2.93 & 12.30 $\pm$ 8.26 \\
    balance & 95.26 $\pm$ 1.02 & 0.0077 $\pm$ 1.3e-3 & 95.20 $\pm$ 1.01 & 8.37 $\pm$ 1.03 \\
    australian  & 85.73 $\pm$ 2.04 & 0.0076 $\pm$ 9.75e-5 & 84.49 $\pm$ 1.18 & 407.97 $\pm$ 167.73 \\
    crx   & 69.56 $\pm$ 2.79 & 0.0095 $\pm$ 1.36e-3 & 67.79 $\pm$ 3.47 & 498.04 $\pm$ 35.22 \\
    transfusion  & 78.19 $\pm$ 3.25 & 0.0082 $\pm$ 8.21e-4 & 77.13 $\pm$ 2.26 & 173.06 $\pm$ 44.12 \\
    tic tac toe & 74.22 $\pm$ 5.50 & 0.038 $\pm$ 4.9e-2 & 73.91 $\pm$ 6.11 & 24.13 $\pm$ 6.81 \\    
    sorlie \tnote{1} & 94.084 $\pm$ 1.54 & 0.165 $\pm$ 0.15 & 90.19 $\pm$ 2.47 & 187.50 $\pm$ 1.37 \\
    secom \tnote{1} & 87.87 $\pm$ 1.88 & 957.00 $\pm$ 87.29 & 86.04 $\pm$ 0.82 & 6359.78 $\pm$ 15.93 \\
    tian \tnote{1} & 81.71 $\pm$ 1.43 & 1.39 $\pm$ 0.67 & 80.92 $\pm$ 1.39 & 7832.76 $\pm$ 6.31 \\
    \bottomrule
    \end{tabular}%
  \begin{tablenotes}
  \item[1] {Marked datasets were run on shared computing platforms, and compute times may not be representative.}
  \end{tablenotes}
  \end{threeparttable}
  \label{table1}%
\end{table}%

Table \ref{table2} summarizes five fold cross validation results of the kernel MCM on a number of datasets. A Gaussian or Radial Basis Function (RBF) kernel was chosen for both the MCM and SVM. The width of the MCM Gaussian kernel and the value of $C$ were chosen by using a grid search.

%

\begin{table}[htbp]
  \centering
\footnotesize\setlength{\tabcolsep}{2.5pt}
  \caption{Kernel MCM Results}
\begin{threeparttable}
\begin{tabular}{c|ccc|ccc}
\hline
	 & & Kernel MCM &  & & Kernel SVM &  \\ \hline
	datasets & accuracy & CPU time (s) & \#SV & accuracy & CPU time (s) & \#SV \\ \hline

    \midrule
          &       &       &       &       &       &  \\
    blogger & 88.00 $\pm$ 4.00 & 0.32 $\pm$ 0.03 & 22.20 $\pm$ 5.91 & 81.00 $\pm$ 10.20 & 2573 $\pm$ 49.2 & 51.20 $\pm$ 3.06 \\
    fertility diagnosis & 89.00 $\pm$ 2.00 & 0.18 $\pm$ 0.09 & 9.80 $\pm$ 19.60 & 88.00 $\pm$ 9.27 & 8.03 $\pm$ 1.95 & 38.20 $\pm$ 1.60 \\
    promoters  & 84.93 $\pm$ 1.56 & 0.45 $\pm$ 0.39 & 82.40 $\pm$ 2.73 & 75.59 $\pm$ 7.63 & 4.40 $\pm$ 1.33 & 83.80 $\pm$ 0.98 \\
    echocardiogram  & 89.34 $\pm$ 4.57 & 0.31 $\pm$ 0.01 & 12.00 $\pm$ 0.00 & 87.14 $\pm$ 7.27 & 8.58 $\pm$ 1.91 & 48.00 $\pm$ 2.10 \\
    teaching assistant & 74.83 $\pm$ 2.60 & 0.39 $\pm$ 0.13 & 26.60 $\pm$ 32.43 & 68.88 $\pm$ 6.48 & 4192 $\pm$ 162 & 86.00  3.22 \\
    hepatitis  & 85.80 $\pm$ 8.31 & 0.44 $\pm$ 0.02 & 20.00 $\pm$ 0.00 & 82.57 $\pm$ 6.32 & 3561 $\pm$ 4392 & 72.20 $\pm$ 4.31 \\
    hayes & 81.82 $\pm$ 7.28 & 0.31 $\pm$ 0.05 & 3.23 $\pm$ 1.11 & 79.57 $\pm$ 6.60 & 1427 $\pm$ 54.7 & 84.20 $\pm$ 2.04 \\
    plrx  & 71.99 $\pm$ 5.81 & 0.41 $\pm$ 0.10 & 4.40 $\pm$ 8.80 & 71.41 $\pm$ 6.04 & 144.21 $\pm$ 5816 & 116.2 $\pm$ 6.14 \\
    seed  & 97.13 $\pm$ 0.95 & 0.79 $\pm$ 0.01 & 11.20 $\pm$ 5.71 & 91.90 $\pm$ 2.86 & 3362 $\pm$ 85.1 & 51.80 $\pm$ 1.72 \\
    glass & 96.23 $\pm$ 2.77 & 1.69 $\pm$ 0.50 & 36.00 $\pm$ 11.49 & 90.64 $\pm$ 5.09 & 20475 $\pm$ 832 & 64.80 $\pm$ 2.40 \\
    heart statlog  & 84.44 $\pm$ 3.21 & 1.32 $\pm$ 0.76 & 10 $\pm$ 2.23 & 83.7 $\pm$ 1.54 & 1547 $\pm$ 324.52 & 124.6 $\pm$ 4.15 \\
    horsecolic  & 82.33 $\pm$ 4.03 & 3.84 $\pm$ 2.31 & 36.60 $\pm$ 17.70 & 81.33 $\pm$ 4.14 & 13267 $\pm$ 2646 & 187.2 $\pm$ 3.27 \\
    haberman  & 73.49 $\pm$ 3.85 & 1.23 $\pm$ 0.32 & 8.50 $\pm$ 7.00 & 72.81 $\pm$ 3.51 & 2087 $\pm$ 750 & 138.2 $\pm$ 3.27 \\
    ecoli & 97.32 $\pm$ 1.73 & 3.47 $\pm$ 0.30 & 24.00 $\pm$ 1.41 & 96.42 $\pm$ 2.92 & 11829 $\pm$ 248 & 57.00 $\pm$ 4.65 \\
    house voters & 95.87 $\pm$ 1.16 & 4.24 $\pm$ 0.83 & 17.80 $\pm$ 8.91 & 95.42 $\pm$ 2.04 & 8827 $\pm$ 349 & 93.60 $\pm$ 3.93 \\
    wholesale customer & 92.72 $\pm$ 1.54 & 7.31 $\pm$ 0.93 & 39.00 $\pm$ 10.64 & 90.90 $\pm$ 1.90 & 9243 $\pm$ 362 & 123.40 $\pm$ 2.15 \\
    IPLD  & 72.03 $\pm$ 3.20 & 4.06 $\pm$ 5.02 & 23.40 $\pm$ 30.50 & 70.15 $\pm$ 2.24 & 9743 $\pm$ 322 & 311.60 $\pm$ 5.31 \\
    balance & 97.64 $\pm$ 1.32 & 8.78 $\pm$ 1.32 & 14.60 $\pm$ 0.49 & 97.60 $\pm$ 0.51 & 15442 $\pm$ 651 & 143.00 $\pm$ 4.23 \\
    australian  & 85.65 $\pm$ 2.77 & 103.45 $\pm$ 18.04 & 108.80 $\pm$ 1.60 & 84.31 $\pm$ 3.01 & 94207 $\pm$ 4476 & 244.8 $\pm$ 4.64 \\
    crx   & 69.56 $\pm$ 2.90 & 5.95 $\pm$ 2.55 & 3.40 $\pm$ 6.80 & 69.27 $\pm$ 2.62 & 19327 $\pm$ 5841 & 404.4 $\pm$ 8.69 \\
    transfusion  & 77.00 $\pm$ 2.84 & 7.08 $\pm$ 0.69 & 6.00 $\pm$ 3.52 & 76.73 $\pm$ 2.88 & 18254 $\pm$ 1531 & 302.20 $\pm$ 7.55 \\
    tic tac toe & 98.32 $\pm$ 0.89 & 12.55 $\pm$ 0.56 & 10.00 $\pm$ 0.00 & 93.94 $\pm$ 2.10 & 18674 $\pm$ 973 & 482.60 $\pm$ 3.93 \\
    sorlie \tnote{1} & 98.82 $\pm$ 2.35 & 0.44 $\pm$ 0.15 & 50 $\pm$ 4.77 & 97.644 $\pm$ 2.88 & 78.63 $\pm$ 9.81 & 68.95 $\pm$ 3.72\\
    secom \tnote{1} & 94.11 $\pm$ 2.23 & 1521 $\pm$ 75.5 & 382.8 $\pm$ 44.23 & 92.29 $\pm$ 0.82 & 38769.25 $\pm$ 8.87 & 593.2 $\pm$ 17.22\\
    tian \tnote{1} & 97.09 $\pm$ 3.83 & 2.05 $\pm$ 0.199 & 70.4 $\pm$ 3.26 & 95.188 $\pm$ 4.26 & 88.97 $\pm$ 3.26 & 75.6 $\pm$ 1.01\\
    \bottomrule
    \end{tabular}%
 \begin{tablenotes}
  \item[1] {Marked datasets were run on shared computing platforms, and compute times may not be representative.}
  \end{tablenotes}
  \end{threeparttable}
    \label{table2}%
\end{table}%

The table shows test set accuracies and the number of support vectors for both the kernel MCM, and the classical SVM with a Gaussian kernel. The results indicate that the kernel MCM yields better generalization than the SVM. In the case of kernel classification, the MCM uses fewer support vectors, generally about one-third the number used by SVMs. In the case of many of the datasets, the MCM uses less than one-tenth the number of support vectors required by a SVM. The code for the MCM classifier would be available from the author's website. The large difference with the SVM results indicates that despite good performance, SVM solutions may still be far from optimal. Vapnik \citep{vapnik95} showed that
\begin{equation}
E(P_{error}) \leq \frac{E(\# \mbox{support vectors})}{\# \mbox{training samples}},
\end{equation}\label{vcbound2}
where $E(P_{error})$ denotes the expected error on test samples taken from the general distribution, $\# \mbox{training samples}$  denotes the number of training samples, and $E(\#$ support vectors $)$ denotes the expected number of support vectors obtained on training sets of the same size. Although the bound was shown for linearly separable datasets, it does indicate that the number of support vectors is also related to the prediction error. An examination of the table indicates that the proposed approach shows a lower test set error, and also uses a smaller number of support vectors. \\

\begin{table}[htbp]
  \centering
  \caption{Values of $h$ for the linear and the kernel MCM}
    \begin{tabular}{ccr}
    \toprule
    datasets & kernel MCM $h$ & linear MCM $h$\\
    \midrule
          &       &  \\
    blogger & 3.73 $\pm$ 1.90 & \multicolumn{1}{c}{2.53 $\pm$ 0.78} \\
    fertility diagnosis & 1.00 $\pm$ 0.00 & \multicolumn{1}{c}{2.72 $\pm$ 1.56} \\
    promoters  & 1.00 $\pm$ 0.00 & \multicolumn{1}{c}{35.77 $\pm$ 18.58} \\
    echocardiogram  & 8.00 $\pm$ 2.94 & \multicolumn{1}{c}{35.76 $\pm$ 18.58} \\
    seed  & 8.32 $\pm$ 4.50 & \multicolumn{1}{c}{9.46 $\pm$ 3.28} \\
    hepatitis  & 3.02 $\pm$ 1.49 & \multicolumn{1}{c}{2.02 $\pm$ 1.32} \\
    teaching assistant & 1.38 $\pm$ 0.47 & \multicolumn{1}{c}{2.12 $\pm$ 0.94} \\\
    plrx  & 1.00 $\pm$ 0.00 & \multicolumn{1}{c}{1.00 $\pm$ 0.00} \\
    hayes & 3.23 $\pm$ 1.11 & \multicolumn{1}{c}{5.00 $\pm$ 0.00} \\
    glass & 8.80 $\pm$ 7.96 & \multicolumn{1}{c}{13.64 $\pm$ 3.32} \\
    heart statlog  & 2.12 $\pm$ 1.87 & \multicolumn{1}{c}{3.79 $\pm$ 1.08} \\
    horsecolic  & 1.20 $\pm$ 0.40 & \multicolumn{1}{c}{1.00 $\pm$ 0.00} \\
    haberman  & 1.18 $\pm$ 0.24 & \multicolumn{1}{c}{1.06 $\pm$ 0.07} \\
    ecoli & 2.41 $\pm$ 1.02 & \multicolumn{1}{c}{4.85 $\pm$ 1.93} \\
    house voters & 4.24 $\pm$ 0.83 & \multicolumn{1}{c}{4.41 $\pm$ 4.21} \\
    wholesale customer & 2.36 $\pm$ 0.97 & \multicolumn{1}{c}{19.79 $\pm$ 6.55} \\
    IPLD  & 6.46 $\pm$ 3.22 & \multicolumn{1}{c}{1.00 $\pm$ 0.00} \\
    balance & 8.73 $\pm$ 7.21 & \multicolumn{1}{c}{16.60 $\pm$ 0.80} \\
    australian  & 1.00 $\pm$ 0.00 & \multicolumn{1}{c}{1.86 $\pm$ 0.43} \\
    crx   & 1.00 $\pm$ 0.00 & \multicolumn{1}{c}{1.00 $\pm$ 0.00} \\
    transfusion  & 2.07 $\pm$ 1.36 & \multicolumn{1}{c}{1.57 $\pm$ 0.71} \\
    tic tac toe & 1.52 $\pm$ 0.06 & \multicolumn{1}{c}{3.07 $\pm$ 1.04} \\
    sorlie & 1.00 $\pm$ 0.00 & \multicolumn{1}{c}{1.12 $\pm$ 0.17} \\
    secom & 8.32 $\pm$ 0.71 & \multicolumn{1}{c}{1.00 $\pm$ 0.00} \\
    tian & 1.25 $\pm$ 0.27 & \multicolumn{1}{c}{5.31 $\pm$ 2.26} \\
     \bottomrule
    \end{tabular}%
  \label{htable}%
\end{table}%

Table \ref{htable} shows the values of $h$ for the linear MCM and the kernel MCM on the benchmark datasets. The values have been indicated as mean $\pm$ standard deviation, computed over the five folds in a five-fold cross validation setting. Note that $h^2$ is an exact bound on the VC dimension $\gamma$, and thus an approximate measure of the capacity of the learning machine. The table indicates that the values of $h$ for the kernel MCM are generally smaller than that for the linear MCM. This also reflects in the better generalization achieved by the kernel MCM.\\



\section{Acknowledgment}
The author would like to thank Prof. Suresh Chandra of the Department of Mathematics, IIT Delhi, for his valuable comments and a critical appraisal of the manuscript. The extensive simulations were the result of the untiring efforts of Siddarth Sabharwal and Sanjit Singh Batra. Early simulations on the linear model were done by Prasoon Goel.

\section{Conclusion} \label{conclusion}
In this paper, we propose a way to build a hyperplane classifier, termed as the Minimal Complexity Machine (MCM), that attempts to minimize a bound on the VC dimension. The classifier can be found by solving a linear programming problem. Experimental results show that the learnt classifier outperforms the classical SVM in terms of generalization accuracies on a number of selected benchmark datasets. At the same time, the number of support vectors is less, often by a substantial factor. It has not escaped our attention that the proposed approach can be extended to least squares classifiers, as well as to tasks such as regression; in fact, a large number of variants of SVMs can be re-examined with the objective of minimizing the VC dimension.

\appendices
\section{An Exact Bound on the VC dimension $\gamma$} \label{appendix1}
We derive an exact or tight ($\Theta$) bound on the VC dimension $\gamma$. Vapnik \citep{vapnik98} showed that the VC dimension $\gamma$ for fat margin hyperplane classifiers with margin $d \geq d_{min}$ satisfies
\begin{equation}
\gamma \leq 1 + \operatorname{Min}(\frac{R^2}{d^2_{min}}, n)
\end{equation}
where $R$ denotes the radius of the smallest sphere enclosing all the training samples. We first consider the case of a linearly separable dataset. By definition, there exists a hyperplane $u^Tx + v = 0$ with positive margin $d$ that can classify these points with zero error. We can always choose $d_{min} = d$; for all further discussion we assume that this is the case. Without loss of generality, we consider hyperplanes passing through the origin. To see that this is possible, we augment the co-ordinates of all samples with an additional dimension or feature whose value is always $1$, i.e. the samples are given by $\hat{x}^i \leftarrow \{x^i; 1\}, i = 1, 2, ..., M$; also, we assume that the weight vector is $(n+1)$-dimensional, i.e. $\hat{u} \leftarrow \{u; v\}$.

Then, the margin, which is the distance of the closest point from the hyperplane, is given by 
\begin{equation}
 d = \operatorname*{Min}_{i = 1, 2, ...M} \frac{\|\hat{u}^T \hat{x}^i\|}{\|\hat{u}\|}
\end{equation}\label{geomargin0}

\begin{equation} 
\frac{R}{d} = \frac{\operatorname*{Max}_{i = 1, 2, ...M}\|\hat{x}^i\|}{\operatorname*{Min}_{i = 1, 2, ...M} \frac{\|\hat{u}^T \hat{x}^i\|}{\|\hat{u}\|}} = \frac{\operatorname*{Max}_{i = 1, 2, ...M} \|\hat{u}\| \| \hat{x}^i\|}{\operatorname*{Min}_{i = 1, 2, ...M} \|\hat{u}^T \hat{x}^i\|} \label{Rbyd1}
\end{equation}

From the Cauchy-Schwarz inequality, we have
\begin{gather}
\|\hat{u}^T \hat{x}^i\| \leq \|\hat{u}\| \|\hat{x}^i\| \label{csineq1}\\
\implies \operatorname*{Max}_{~i = 1, 2, ..., M} \|\hat{u}^T \hat{x}^i\| \leq \operatorname*{Max}_{~i = 1, 2, ..., M} \|\hat{u}\| \|\hat{x}^i\| \label{csineq2}
\end{gather}

Therefore, from (\ref{Rbyd1}), we have
\begin{gather}
 \frac{\operatorname*{Max}_{i = 1, 2, ..., M} \|\hat{u}^T \hat{x}^i\|}{\operatorname*{Min}_{i = 1, 2, ..., M} \|\hat{u}^T \hat{x}^i\|} ~ \leq \frac{R}{d} \label{csineq3}
\end{gather}
or, in terms of the original variables $u$ and $v$,

\begin{gather}\label{bound1a}
 \frac{\operatorname*{Max}_{i = 1, 2, ..., M} \|u^T x^i + v\|}{\operatorname*{Min}_{i = 1, 2, ..., M} \|u^T x^i + v\|} \leq \frac{R}{d}
\end{gather}\\

Denoting
\begin{gather}
 h = \frac{\operatorname*{Max}_{i = 1, 2, ..., M} \|u^T x^i + v\|}{\operatorname*{Min}_{i = 1, 2, ..., M} \|u^T x^i + v\|},
\end{gather}
we can write
\begin{gather}\label{bound1}
 h \leq \frac{R}{d}, \\
 \implies h^2 \leq (\frac{R}{d})^2 < 1 + (\frac{R}{d})^2.
\end{gather}
Assuming that the dimension $n$ of the data samples is sufficiently large, we have, from (\ref{eqnh}),
\begin{gather}
 \gamma \leq 1 + (\frac{R}{d})^2.
\end{gather}
Hence, $\exists \beta \in \Re$,  $\beta > 0$, such that
\begin{gather}\label{gammaleqh2}
\gamma \leq \beta h^2.
\end{gather}

We also note that
\begin{equation}\label{minh}
 h^2 \geq 1,
\end{equation}
the minimum being achieved when all samples are equidistant from the separating hyperplane.

Also note that the VC dimension $\gamma$ satisfies
\begin{equation}\label{minvc}
 \gamma \geq 1.
\end{equation}

In short, both $h^2$ and $\gamma$ have the same lower and upper bounds, i.e. they are of the same order. Therefore, from (\ref{gammaleqh2}), (\ref{minh}) and (\ref{minvc}), we note that there exist constants $\alpha, \beta > 0$, $\alpha, \beta \in \Re$ such that

\begin{equation}\label{thetavc2}
 \alpha h^2 \leq \gamma \leq \beta h^2,
\end{equation}
or, in other words, $h^2$ constitutes a tight or exact ($\theta$) bound on the VC dimension $\gamma$.\\

Since the dataset has been assumed to be linearly separable, we have
\begin{eqnarray}
 u^Tx^i + v \geq 0, & \mbox{if } y_i = 1 \mbox{   (Class 1 points)} \label{moduvg0}\\
 u^Tx^i + v \leq 0, & \mbox{if } y_i = -1 \mbox{  (Class -1 points)} \label{moduvl0}
\end{eqnarray}

We also have
\begin{equation}\label{moduv}
 \|u^Tx^i + v\| = \begin{cases} u^Tx^i + v, & \mbox{if } u^Tx^i + v \geq 0 \\ 
                                -(u^Tx^i + v), & \mbox{if } u^Tx^i + v \leq 0 \end{cases}
\end{equation}

Therefore, from (\ref{moduvg0}), (\ref{moduvl0}), and (\ref{moduv}), we can write
\begin{equation}
 \|u^Tx^i + v\| = y_i \cdot [u^T x^i + v], ~i = 1, 2, ..., M.
\end{equation}

Therefore,
\begin{gather}\label{eqhrepeat}
 h = \frac{\operatorname*{Max}_{i = 1, 2, ..., M} \;y_i(u^T x^i + v)}{\operatorname*{Min}_{i = 1, 2, ..., M} \;y_i(u^T x^i + v)}.
\end{gather}

\section{The Hard Margin MCM Formulation} \label{appendix2}
In this appendix, we derive the hard margin MCM formulation in the input space. We begin with the optimization problem in (\ref{minh1}), which was obtained from the exact bound on $\gamma$ derived in Appendix \ref{appendix1}. In deriving the exact bound in Appendix \ref{appendix1}, we assumed that the separating hyperplane $u^Tx + v = 0$ correctly separates the linearly separable training points; hence, no other constraints are present in the optimization problem (\ref{minh1}). For the convenience of the reader, (\ref{minh1}) [also (\ref{eqhrepeat})] is reproduced below.
\begin{equation}
\operatorname*{Minimize  }_{u, v} \; h ~=~ \frac{\operatorname*{Max}_{i = 1, 2, ..., M} \; y_i(u^T x^i + v)}{\operatorname*{Min}_{i = 1, 2, ..., M} \; y_i(u^T x^i + v)} \nonumber
\end{equation}
The problem (\ref{minh1}) may be rewritten as

\begin{gather}
\operatorname*{Min}_{u, v, g, l} ~~\frac{g}{l} \label{obj1}\\
g \geq y_i \cdot [{u^T x^i + v}], ~i = 1, 2, ..., M \label{cons1}\\
l \leq y_i \cdot [{u^T x^i + v}], ~i = 1, 2, ..., M \label{cons2}
\end{gather}

This is a linear fractional programming problem \citep{numoptbook}. We apply the Charnes-Cooper transformation \citep[see][p. 463]{numoptbook}. This consists of introducing a variable $p = \frac{1}{l}$, which we substitute into (\ref{obj1})-(\ref{cons2}) to obtain

\begin{gather}
\operatorname*{Min}_{u, v, g, p, l} ~~h = g \cdot p \label{obj2}\\
g \cdot p \geq y_i \cdot [p \cdot {u^T x^i + p \cdot v}], ~i = 1, 2, ..., M \label{cons21}\\
l \cdot p \leq y_i \cdot [p \cdot {u^T x^i + p \cdot v}], ~i = 1, 2, ..., M \label{cons22}\\
p \cdot l = 1 \label{cons24}
\end{gather}

Denoting $w \equiv p \cdot u$, $b \equiv p \cdot v$, and noting that $p \cdot l = 1$, we obtain the following optmization problem.
\begin{gather}
\operatorname*{Min}_{w, b, h} ~~h \label{obj3}\\
h \geq y_i \cdot [{w^T x^i + b}], ~i = 1, 2, ..., M \label{cons31}\\
1 \leq y_i \cdot [{w^T x^i + b}], ~i = 1, 2, ..., M \label{cons32}
\end{gather}

which may be written as
\begin{gather}
\operatorname*{Min}_{w, b, h} ~~h \label{obj4}\\
h \geq y_i \cdot [{w^T x^i + b}], ~i = 1, 2, ..., M \label{cons41}\\
y_i \cdot [{w^T x^i + b}] \geq 1, ~i = 1, 2, ..., M \label{cons42}
\end{gather}

We refer to the problem (\ref{obj4}) - (\ref{cons42}) as the hard margin Linear Minimum Complexity Machine (Linear MCM). Note that $h^2$ is an exact bound on $\gamma$, the VC dimension of the classifier.

\bibliographystyle{elsarticle-num}
\bibliography{linear-struct-min3-els-arxiv}

\end{document}